\documentclass[letterpaper, conference]{IEEEtran}

\IEEEoverridecommandlockouts                              

\usepackage{subcaption}
\usepackage{hyperref}       
\usepackage{booktabs}       
\usepackage{placeins}
\usepackage[numbers, sort&compress]{natbib}
\usepackage{graphicx}
\usepackage[acronym,shortcuts]{glossaries
}
\usepackage{comment}
\usepackage{enumerate}
\usepackage[capitalise]{cleveref}
\usepackage[table,xcdraw]{xcolor}
\usepackage{siunitx}
\usepackage{bm} 
\usepackage{amssymb} 
\usepackage{amsmath}
\usepackage[english]{babel}
\usepackage{multirow}
\usepackage{pifont}
\usepackage{stfloats}
\usepackage{makecell}
\usepackage{array}
\hypersetup{hidelinks}

\definecolor{lightgray}{gray}{0.9}

\def\BibTeX{{\rm B\kern-.05em{\sc i\kern-.025em b}\kern-.08em
    T\kern-.1667em\lower.7ex\hbox{E}\kern-.125emX}}
\title{\LARGE \bf
MAP2: Model- and Acceleration-Based Pursuit with MPC and Gaussian Process Residual Learning for Autonomous Racing
}

\author{Weizhan Huang\IEEEauthorrefmark{2}, Cheng Hu\IEEEauthorrefmark{2}, Edoardo Ghignone\IEEEauthorrefmark{3}, Nicolas Baumann\IEEEauthorrefmark{3}, Yiqin Wang\IEEEauthorrefmark{2}, Hongye Su\IEEEauthorrefmark{2}, \\ Lei Xie\IEEEauthorrefmark{2}, and Michele Magno\IEEEauthorrefmark{3} \\
\thanks{\IEEEauthorrefmark{2}Weizhan Huang, Cheng Hu, Lei Xie, Hongye Su and Yiqin Wang are associated with Zhejiang University.}%
\thanks{\IEEEauthorrefmark{3}Nicolas Baumann, Edoardo Ghignone, and Michele Magno are associated with the Center for Project-Based Learning, D-ITET, ETH Zurich.
\emph{(Weizhan Huang, Cheng Hu, Edoardo Ghignone, Nicolas Baumann and Yiqin Wang contributed equally to this work.) (Corresponding author: Cheng Hu.)}}}%

\begin{document}
\setlength{\abovedisplayskip}{6pt plus 2pt minus 2pt}
\setlength{\belowdisplayskip}{6pt plus 2pt minus 2pt}
\setlength{\abovedisplayshortskip}{3pt plus 1pt minus 1pt}
\setlength{\belowdisplayshortskip}{3pt plus 1pt minus 1pt}
\newacronym{lut}{LUT}{Lookup Table}
\newacronym{VESC}{VESC}{Vedder Electronic Speed Controller}
\newacronym{ALT}{ALT}{average lap time}
\newacronym{FLT}{FLT}{fastest lap time}
\newacronym{ALE}{ALE}{average lateral error}
\newacronym{MLE}{MLE}{maximum lateral error}
\newacronym{mpc}{MPC}{Model Predictive Control}
\newacronym{mpcc}{MPCC}{Model Predictive Contouring Controller}
\newacronym{rl}{RL}{Reinforcement Learning}
\newacronym{mlp}{MLP}{Multilayer Perceptron}
\newacronym{forl}{FoRL}{Foundations of Reinforcement Learning}
\newacronym{ml}{ML}{Machine Learning}
\newacronym{sac}{SAC}{Soft Actor Critic}
\newacronym{ppo}{PPO}{Proximal Policy Optimization}
\newacronym{ai}{AI}{Artificial Intelligence}
\newacronym{nn}{NN}{Neural Network}
\newacronym{sota}{SotA}{State-of-the-Art}
\newacronym{esc}{ESC}{Electronic Speed Controller}
\newacronym{ros}{ROS}{Robot Operating System}
\newacronym{imu}{IMU}{Inertial Measurement Unit}
\newacronym{ekf}{EKF}{Extended Kalman Filter}
\newacronym{slam}{SLAM}{Simultaneous Localization And Mapping}
\newacronym{ads}{ADS}{Autonomous Driving Systems}
\newacronym{obc}{OBC}{On Board Computer}
\newacronym{qp}{QP}{Quadratic Programming}
\newacronym{uav}{UAV}{Unmanned Aerial Vehicles}
\newacronym{cg}{CG}{Center of Gravity}
\newacronym{em}{EM}{Expectation Maximization}
\newacronym{rms}{RMS}{Root Mean Square}
\newacronym{map}{MAP}{Model- and Acceleration-based Pursuit}
\newacronym{pd}{PD}{Proportional-Derivative}
\newacronym{iac}{IAC}{Indy Autonomous Challenge}
\newacronym{fsd}{FSD}{Formula Student Driverless}
\newacronym{map2}{MAP2}{Model- and Acceleration-Based Pursuit with MPC and Gaussian Process Residual Learning}
\newacronym{pp}{PP}{Pure Pursuit}
\newacronym{cpu}{CPU}{Central Processing Unit}
\newacronym{awd}{AWD}{All-Wheel Drive}
\newacronym{rwd}{RWD}{Rear-Wheel Drive}
\newacronym{sgp}{SGP}{Sparse Gaussian Process}
\newacronym{gp}{GP}{Gaussian Process}

\maketitle
\thispagestyle{empty}
\pagestyle{empty}

\begin{abstract}
Autonomous racing requires accurate trajectory tracking near the handling limits of a vehicle while maintaining low computational latency. Geometric controllers are computationally efficient, but the Ackermann steering geometry becomes invalid under limit-handling conditions. \gls{map} preserves the simplicity of geometric approaches while leveraging tire dynamics. Yet MAP remains fundamentally a geometric controller that relies on Ackermann steering geometry, and its tire model may not fully capture the vehicle’s actual dynamic response. This paper presents MAP2, a model-based pursuit controller that combines a curvature-based kinematic MPC and a \gls{sgp} residual correction. The proposed algorithm uses MPC to optimize kinematic control inputs over a prediction horizon and maps them to steering commands through a tire dynamics model augmented with \gls{sgp} residual correction. Real-world vehicle experiments demonstrate substantial reductions in lateral tracking error and lap time. Compared with \gls{map} and \gls{pp}, MAP2 reduces the average lateral tracking error by 37.99\% and 44.65\%, respectively, while reducing average lap time by at least 1.5\%.
\end{abstract}

\section{INTRODUCTION}
Autonomous racing challenges autonomous driving systems to achieve accurate, low-latency trajectory tracking near the vehicle’s handling limits. These challenges, prevalent in racing scenarios, also offer valuable insights for standard driving situations \cite{indyautonomous}. Academic competitions, from full-scale events such as \gls{fsd} \cite{amz_fullstack} and \gls{iac} \cite{indyautonomous} to F1TENTH-scale racing \cite{f110, forzaeth}, provide testbeds for studying these challenges. Though these competitions vary, their common goal is to complete laps in the shortest time \cite{amz_fullstack, indyautonomous, forzaeth}.

This goal introduces a complex control challenge, pushing \gls{ads} to operate at their physical limits. The highly nonlinear behavior of racecars, dominated by vehicle and tire dynamics at the limits of friction \cite{liniger_mpcc,map,forzaeth}, makes precise trajectory tracking at these limits the paramount control task in autonomous racing \cite{map}.

Traditionally, \gls{mpc} is widely used for high-performance autonomous racing \cite{liniger_mpcc, mpccurv}, offering predictive control but depending on accurate vehicle models and incurring increasing computational cost as model complexity grows. Consequently, while \gls{mpc} theoretically delivers an optimal control solution for autonomous racing, practical limitations such as state estimation inaccuracies and model mismatches often curtail its effectiveness \cite{map}, complicating its deployment in a dynamic racing environment. Hence, an alternative to \gls{mpc} controllers, \gls{map} \cite{map}, has been introduced, which integrates a geometric controller with the complex tire dynamics, through a \gls{lut}-based approach. However, the geometric nature of its underlying \gls{pp} \cite{pp} and the mismatch between the offline \gls{lut} and the real vehicle dynamics still leave a large margin for tracking performance.

Therefore, in this paper, we propose MAP2, a computationally lightweight controller architecture that builds upon the proven capabilities of the \gls{map} controller \cite{map, forzaeth}. MAP2 significantly enhances tracking performance by integrating a highly efficient \gls{mpc} that adopts the model-based approach of the original \gls{map}, and a residual \gls{sgp} on top of the \gls{lut} baseline. This integration allows MAP2 to achieve constraint satisfaction and execute accurate steering commands over a receding horizon while preserving the simplicity and parameter robustness of \gls{map}. We have experimentally validated the MAP2 controller on the computationally constrained 1:10-scale F1TENTH platform, comparing it against \gls{sota} autonomous racing controllers such as \gls{map} \cite{map} and the traditional \gls{pp} controller. The main contributions of this work are as follows:
\begin{enumerate}[I]
\item \textbf{MPC-Based Predictive Lateral-Acceleration Generation:} A Frenet-frame kinematic \gls{mpc} is introduced to replace the geometric look-ahead based lateral acceleration calculation in \gls{map}. Instead of deriving the lateral acceleration command $a_c$ from the look-ahead distance $L_d$, the proposed \gls{mpc} optimizes virtual inputs over a prediction horizon then convert them into $a_c$. This formulation enables predictive lateral acceleration planning for operation near the vehicle's handling limits.

\item \textbf{SGP-Based Compensation for Model Mismatch:} An \gls{sgp} residual model is learned from real-vehicle data to compensate for this mismatch by correcting the \gls{lut}-based steering commands. The sparse formulation reduces the computational cost of SGP inference to 2.0~ms, enabling real-time residual correction.

\item \textbf{Real-world Experimental Validation and Open Source:} MAP2 is validated on a computationally constrained 1:10-scale F1TENTH platform through comparisons with \gls{map} and \gls{pp}, as well as an ablation study without \gls{sgp} residual correction. The experiments demonstrate reduced lateral tracking errors and lap times, together with robustness to tire-parameter modeling errors. To support reproducibility and further research, we release our implementation at \url{https://anonymous.4open.science/r/MAP2-1F88/}.
\end{enumerate}

\section{Related Work}

\gls{mpc} has become one of the dominant approaches for high-performance autonomous racing. By repeatedly solving a finite-horizon optimization problem, \gls{mpc} predicts the future vehicle motion and computes control inputs. Dynamic \gls{mpc} and \gls{mpcc} have demonstrated strong performance on autonomous race cars because they can jointly optimize trajectory tracking and vehicle speed~\cite{liniger_mpcc,mpccurv,amz_fullstack}. However, this performance depends on accurate vehicle and tire models. Near the handling limits, tire forces are highly nonlinear. Modeling these effects accurately is difficult, and discrepancies between the identified model and the real vehicle can directly degrade the control performance~\cite{pacejka1992magic}. GP-based residual learning provides one way to address this limitation. Hewing et al. augment a nominal dynamics model with a GP that learns model errors from data and accounts for the remaining uncertainty within \gls{mpc}~\cite{hewing2020gpcorruction}. Sparse GP formulations have likewise been used to compensate for model mismatch online in safety-critical vehicle control~\cite{hu2025sgp,fsdp}. Nevertheless, online optimization and GP prediction introduce computational demands that must be considered on embedded racing platforms~\cite{forzaeth,small_comp2, map}.

Geometric controllers provide a computationally efficient alternative. These methods derive the steering command from the geometric relationship between the vehicle and a reference trajectory, and therefore do not require a detailed dynamic model. \gls{pp} is a representative example: it selects a reference point ahead of the vehicle at a look-ahead distance \(L_d\) and computes the steering angle using an Ackermann kinematic relationship~\cite{pp}. Because of its low computational cost and simple implementation, such a controller can achieve good tracking performance at low or moderate speeds after proper parameter tuning. However, its geometric assumptions become less accurate as the vehicle approaches the handling limits. At high speed, the lateral dynamics become more nonlinear, tire slip becomes significant, and the curvature of the track may require steering inputs close to the available tire-force and steering limits. The look-ahead distance also introduces an inherent trade-off: a short \(L_d\) can cause oscillatory steering, whereas a long \(L_d\) can increase the tracking error and cause corner cutting. Speed-dependent or otherwise adaptive look-ahead strategies can alleviate this trade-off, but they still do not explicitly describe the nonlinear relationship between tire slip, lateral acceleration, and steering angle.

\gls{map} introduces vehicle dynamics into this lightweight control structure by using a tire model to construct a \gls{lut} that maps the desired lateral acceleration and vehicle velocity to the steering angle~\cite{map}. Compared with a purely geometric \gls{pp} controller, this mapping allows the controller to account for tire slip and the velocity-dependent relationship between lateral acceleration and steering. As a result, \gls{map} retains the low online computational cost of a pursuit controller while providing more accurate lateral control in high-speed and nonlinear operating conditions. Nevertheless, \gls{map} still calculates the desired lateral acceleration from target-point geometry and \(L_d\), without optimizing vehicle motion over a prediction horizon. Its steering commands also remain sensitive to model approximations and parameter identification errors in the \gls{lut}.

These limitations motivate a hybrid design that keeps the online pipeline of a geometric controller while adding short-horizon prediction and data-driven compensation. Unlike~\cite{hewing2020gpcorruction}, which augments a dynamic model with a GP inside an \gls{mpc}, this work applies an \gls{sgp} residual learning to the \gls{lut}-based steering mapping, so that the nominal controller remains lightweight. The \gls{sgp} uses a small set of inducing points to summarize information from the training data, reducing the inference costs to support real-time compensation~(see Table~\ref{tab:comparison}).

\begin{table}[ht]
\centering
\caption{Comparison of representative autonomous racing controllers}
\label{tab:comparison}
\begin{tabular}{
  >{\centering\arraybackslash}m{2.1cm}|
  >{\centering\arraybackslash}m{1.3cm}|
  >{\centering\arraybackslash}m{1.8cm}|
  >{\centering\arraybackslash}m{1.6cm}}
\toprule
\makecell{\textbf{Controller}}
& \makecell{\textbf{Compute}}
& \makecell{\textbf{High-speed}\\\textbf{tracking$\dagger$}}
& \makecell{\textbf{Model}\\\textbf{sensitivity}} \\
\midrule
Pure Pursuit (PP)    & Low    & Limited  & Low  \\
MPC/MPCC             & High   & High     & High \\
MAP                  & Low    & Improved & High \\
\textbf{MAP2 (ours)} & Medium & High     & Low  \\
\bottomrule
\end{tabular}
\par\smallskip
{\raggedright\small $\dagger$\ It evaluates the lateral tracking error and lap time over the high-speed condition, where substantial tire slip occurs.\par}
\end{table}

\section{Methodology}
\begin{figure}[htbp]
  \centering
  \includegraphics[scale=0.185]{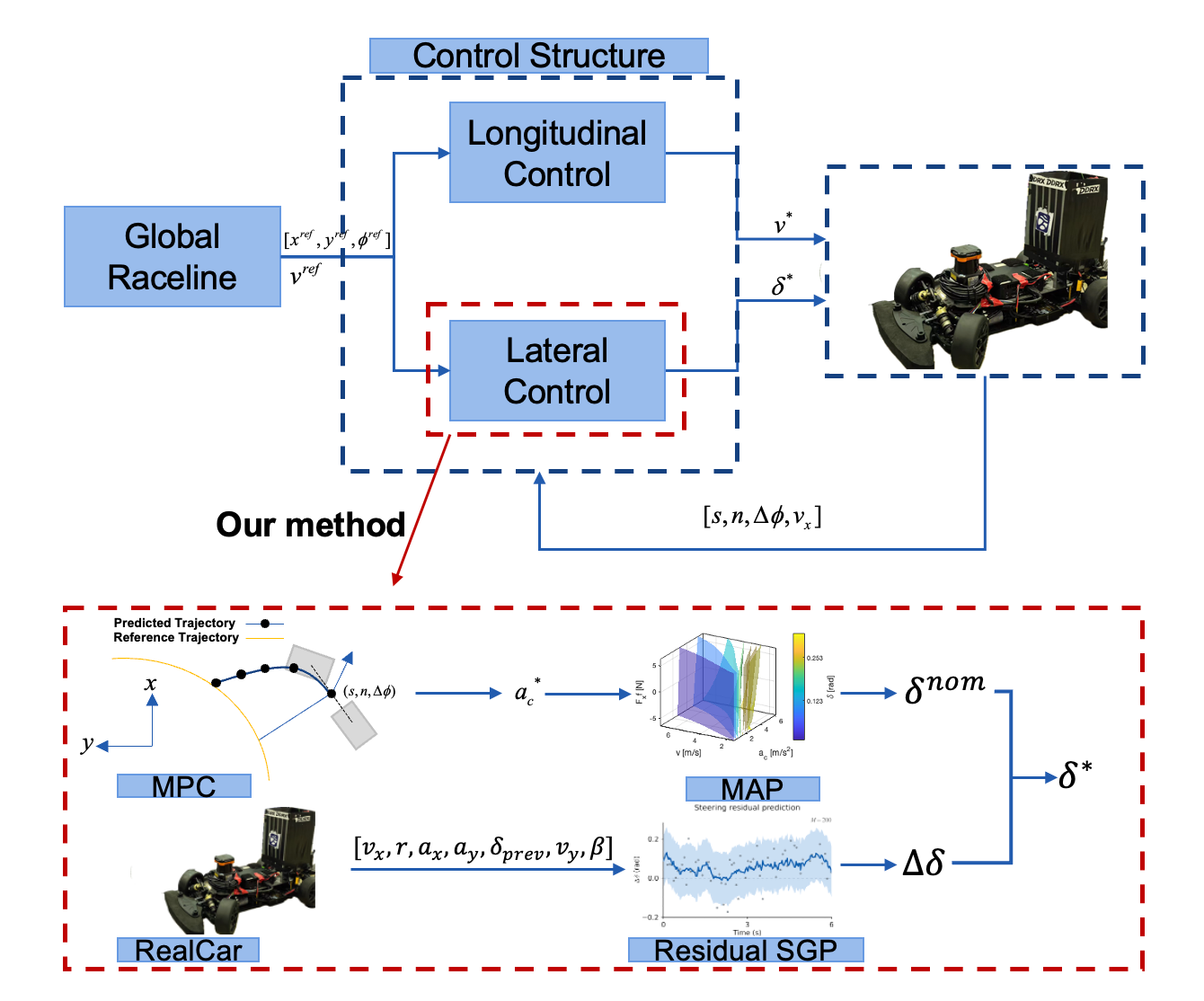}
  \caption{Control structure of the racing car.
}
  \label{con_stru}
\end{figure}

A method of minimum curvature path optimization ~\cite{raceline} is used to generate a global trajectory and reference velocities. The control framework of autonomous racing is divided into longitudinal control and lateral control, as illustrated in \cref{con_stru}. The longitudinal controller computes the velocity input to the car, which is then managed by the lower-level PID controller to control the motors ~\cite{forzaeth}. This paper concentrates on the lateral control of the vehicle. The lateral controller comprises three main components.

First, a Frenet-frame kinematic MPC is designed to compute the optimal virtual longitudinal velocity, lateral velocity and yaw rate for trajectory tracking. The first-stage optimal virtual input is then converted into a lateral acceleration request, and the first-stage optimal longitudinal velocity is returned as the velocity command. Second, a model- and acceleration-based pursuit controller is developed, which adopts an AWD vehicle by using an estimated front-axle longitudinal force. Third, an SGP estimates the residual steering correction associated with the difference between the nominal lookup table and the actual vehicle dynamic response.

\subsection{Curvature Model Predictive Control} 
The Frenet kinematic model is used as a simple representation of the vehicle dynamics, excluding the tire forces. Additionally, by transforming the model from the Cartesian coordinate system to the Frenet coordinate system, it is more convenient to construct the cost function and constraints of the MPC. The motion of the model can be expressed as follows:

\begin{align}
\dot s &=\frac{v_x \text{cos}(e_{\psi}) -v_y \text{sin}(e_{\psi})}{1-\kappa_{r}(s)\cdot n} \\
\dot n &= v_x \text{sin}(e_{\psi})+v_y \text{cos}(e_\psi) \\
 \dot e_\psi &= \dot \Psi-\kappa_{r}(s) \cdot \frac{v_x \text{cos}(e_\psi) -v_y \text{sin}(e_\psi)}{1-\kappa_{r}(s)\cdot n}
\end{align}
where $\kappa_{r}$ indicates the curvature of the reference trajectory. $\Psi$ denotes the vehicle yaw (heading) angle, and $\dot\Psi$ denotes its yaw rate. $s$ denotes the distance traveled along the reference trajectory, $n$ is the lateral deviation from this trajectory. $e_\psi$ represents the heading angle error. The last three variables form the state vector:
\begin{align}
x=\begin{bmatrix} s &n &e_\psi \end{bmatrix}^{T}
\end{align}

The input variables include the longitudinal velocity $v_x$, the lateral velocity $v_y$, and the yaw rate $\dot \Psi$:
\begin{align}u=\begin{bmatrix} v_x &v_y &\dot \Psi \end{bmatrix}^{T}\end{align}

At high speeds, the Ackermann steering assumption no longer holds, and pure-pursuit controllers cannot guarantee that the vehicle remains within the road boundaries; MPC, by contrast, continuously optimizes within the vehicle-model and road-boundary constraints. After discretizing the model (1)–(3) with the Runge--Kutta method at a sampling time of $\Delta t=0.05\,\mathrm{s}$, the optimization problem is formulated as
\begin{align} 
\mathop  { \text{min} }\limits_{u_{k+i|k}, \epsilon} \sum^{N-1}_{i=0}& 
[q_n \cdot n_{k+i|k}^2+q_{e_\psi} \cdot e_{\psi, k+i|k}^2 \notag \\&
+\Vert u_{k+i|k} - u_{k+i|k}^{\mathrm{ref}} \Vert^2_R] \notag \\&
+ p_n \cdot n_{k+N|k}^2+p_{e_\psi} \cdot e_{\psi, k+N|k}^2+\rho \cdot \epsilon^2 \\
s.t. \notag \\
&{x}_{k|k}=x_0\\& {x}_{k+i+1|k}=f(x_{k+i|k}, u_{k+i|k})\\
&\underline{n}-\epsilon \leq n_{k+i|k} \leq \overline{n}+\epsilon \\
&u_{k+i|k}\in \mathcal{U}
\end{align}
where $N$ is the prediction horizon and $R=\mathrm{diag}(q_v,r_{v_y},r_{\dot\Psi})$, with $q_n$, $q_{e_\psi}$, $q_v$, $r_{v_y}$, $r_{\dot\Psi}$, $p_n$, $p_{e_\psi}$, and $\rho$ as weighting parameters. The notation $(\cdot)_{k+i|k}$ denotes the quantity predicted for stage $k+i$ from the information available at control step $k$, where $k$ is the current control step and $i\in\{0,\ldots,N-1\}$ is the prediction-stage index. $x_0$ is the measured/estimated state at the current step, $f(\cdot)$ is the discretized Frenet-frame kinematic model (1)–(3), $\underline{n}$ and $\overline{n}$ are the lower and upper lateral bounds imposed by the track boundaries, and $\mathcal{U}$ is the admissible input set (Table~\ref{tab:ctrl_param}). The reference input is $u^{\mathrm{ref}}_{k+i|k}=[v^{\mathrm{ref}}_{x,k+i|k},\,0,\,\kappa^{\mathrm{ref}}_{k+i|k}v^{\mathrm{ref}}_{k+i|k}]^{\mathsf T}$, where $\kappa^{\mathrm{ref}}$ and $v^{\mathrm{ref}}$ are sampled from the reference trajectory. The slack factor $\epsilon$ softens the lateral constraints and thus widens the feasible set.

Then only the first stage of the input for the MPC solution is used to calculate the optimal lateral acceleration request $a_c^{*}$:
\begin{align}
\label{ac_mpc_eq}
u^{*}_{k|k}=\begin{bmatrix} v_x^* &v_y^* &\dot \Psi^{*} \end{bmatrix}^{T}\\
a_c^{*}=\sqrt{(v_x^*)^2+(v_y^*)^2} \cdot {\dot \Psi^{*}}
\end{align}
where $u^{*}_{k|k}$ is the first-stage optimal input and $(\cdot)^{*}$ denotes the optimal solution.
\subsection{Model- and Acceleration-based Pursuit}
The pure pursuit controller is easily implemented due to its straightforward and efficient formulation:
\begin{align}
\label{pp}
\delta=\text{tan}^{-1}\left(\frac{2\text{sin}(\eta)L_{wb}}{L_d}\right)
\end{align}
where $\delta$ denotes the front-wheel steering angle (steering command), $L_d$ is the look-ahead distance and $L_{wb}$ is the wheelbase. $\eta$ is the angle between the velocity vector and look-ahead point.

We can analyze the relationship between lateral acceleration and the steering angle. The guidance logic in ~\cite{park2004new} denotes the relationship between the lateral acceleration and the look-ahead distance:
\begin{align}
\label{ac_eq}
a_c=v_t \cdot \dot\Psi =2\frac{v_t^2}{L_d}\text{sin}(\eta)
\end{align}
where $a_c$ denotes the requested lateral acceleration and $v_t$ is the tangential speed. Under the assumption of Ackermann steering, the pure pursuit does not consider sideslip; therefore, the longitudinal velocity is equal to the tangential velocity $v_x=v_t$.
Substituting Equation~(\ref{ac_eq}) into Equation~(\ref{pp}) yields:
\begin{align}
\delta=\text{tan}^{-1}\left(\frac{a_c \cdot L_{wb}}{v_x^2}\right)
\end{align}

However, this approach has its drawbacks. When a vehicle makes a sharp turn, it is prone to sideslip. This means that the lateral acceleration input to the pure pursuit is not accurate. Inspired by the mapping relationship between lateral acceleration and the steering angle from pure pursuit, the \gls{map} controller is proposed. The basic principle of \gls{map} is using the vehicle's lateral dynamics to establish a new mapping relationship. 

For an all-wheel-drive vehicle, the front axle also generates longitudinal driving force, which changes the combined tire-force distribution and affects the achievable lateral acceleration. Therefore, in this work, the MAP controller is extended to an all-wheel-drive vehicle by introducing the front-axle longitudinal force as an additional input to the lookup table. The new mapping is established using the lateral dynamics of the \gls{awd} vehicle. The dynamic motions of the lateral velocity \(v_y\) and the yaw rate \(\dot{\Psi}\) are expressed as follows:

\begin{align}
\dot v_y = \frac{1}{m} \left [F_{y,r} + F_{y,f}\cdot \cos\delta + F_{x,f}\cdot \sin\delta\right ] - v_x\cdot \dot \Psi \\
\ddot \Psi = \frac{1}{I_z} \left [-l_r F_{y,r} + l_f (F_{y,f}\cdot \cos\delta + F_{x,f}\cdot \sin\delta)\right ]
\end{align}
where \( m \) is the mass of the vehicle model, and \( I_z \) is the moment of inertia about the yaw axis. The distances from the center of gravity (CoG) to the front and rear wheels are denoted by \( l_f \) and \( l_r \), respectively. Additionally, \( F_{y,f} \) and \( F_{y,r} \) represent the lateral forces on the front and rear tires, \( F_{x,f}\) is the longitudinal force on the front axle.

The lateral tire forces are modeled using the Pacejka formula~\cite{pacejka1992magic}, as shown below:
\begin{align}
F_{y,i}=&\mu F_{z,i}D_i \text{sin}  [C_i\cdot  \notag\\ &\text{arctan}\left(B_i \alpha_i-E_i \left(B_i \alpha_i-\text{arctan}\left(B_i \alpha_i\right) \right) \right) ]
\end{align}
where $i\in\{f,r\}$ denotes the front or rear axle, $\mu$ denotes the coefficient of friction and $F_{z,i}$ represents the load on the tire. The parameters $B_i$, $D_i$, $C_i$, and $E_i$ are tire-specific constants that must be identified. The variable $\alpha_i$ indicates the tire sideslip angle~\cite{map}.

The longitudinal tire force of front-axle is calculated by longitudinal acceleration:
\begin{align}
F_{x,f} = \lambda_f \cdot m \cdot a_x 
\end{align}
where $\lambda_f$ denotes the front-axle drive-force split.

\begin{figure}[htbp]
  \centering
  \includegraphics[scale=0.055]{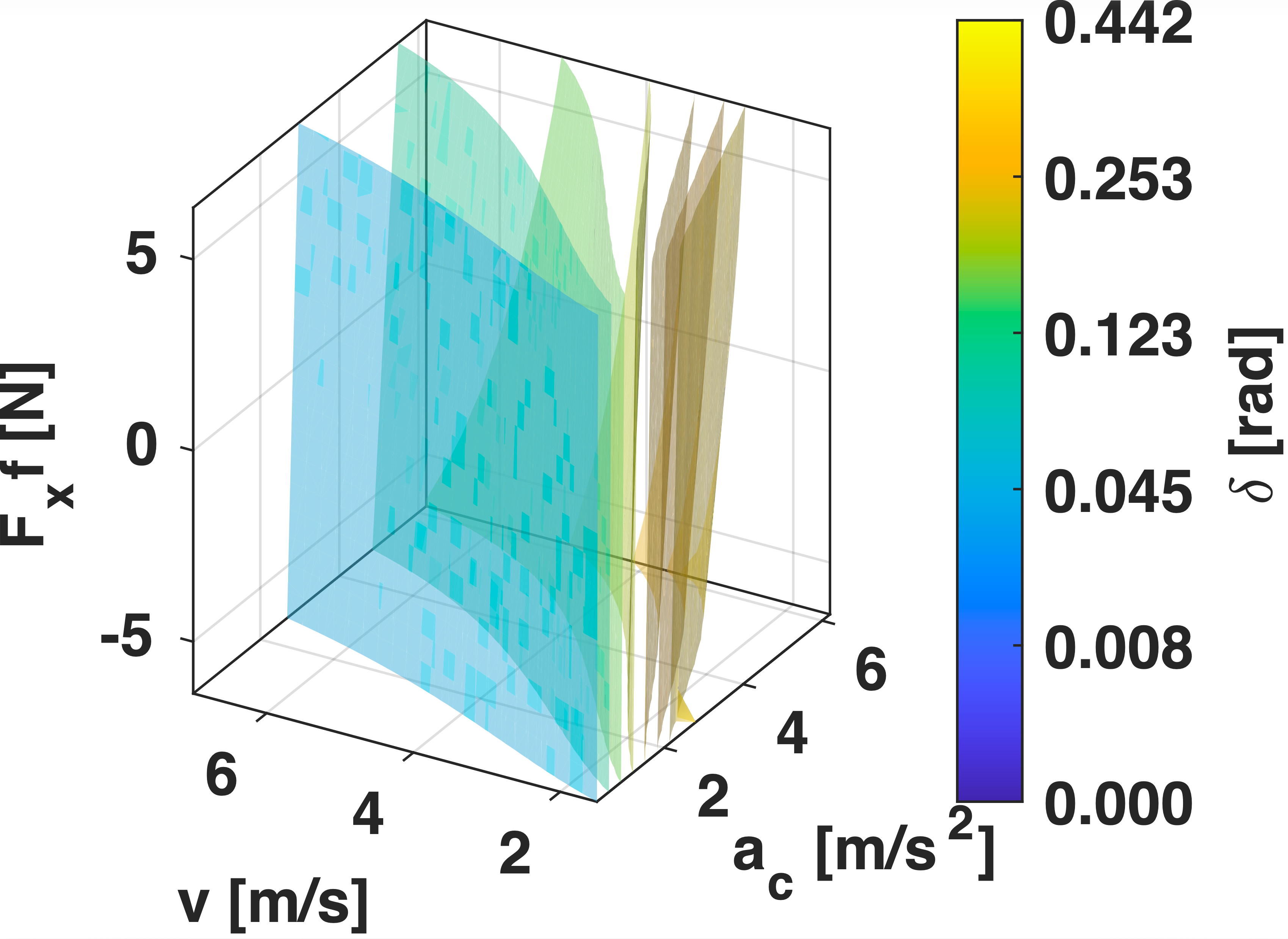}
  \caption{Visualization of the lookup table generated from the
identified car model. From an implementation perspective, the desired acceleration is obtained by interpolating between the closest elements in the table.
}
  \label{fig:LUT}
\end{figure}

Given fixed steering angles, front-axle longitudinal force and longitudinal velocities, a forward simulation of the lateral dynamics is performed. If the dynamics of \(v_y\) and \(\dot{\Psi}\) converge to steady states, this indicates that the vehicle is turning at a fixed radius with a constant lateral acceleration. In racing, high sideslip angles are typically undesirable, so only steady-state points are retained. Steady lateral acceleration is calculated using Equation~(\ref{ac_eq}). A \gls{lut} is used to store the mapping between front wheel steering angles, lateral acceleration, front-axle longitudinal force and longitudinal velocity, as illustrated in  \cref{fig:LUT}.

With the \gls{lut}, the controller is able to retrieve the desired steering angle for the optimal lateral acceleration obtained from curvature MPC, an estimated front-axle longitudinal force and a certain velocity. The final control input can be expressed as follows:
\begin{align}
\delta_{\mathrm{nom, k}}=\text{LUT}(a_c^{*},v_x,F_{x,f})
\end{align}
\subsection{Sparse Gaussian Process Residual Correction}

Due to the mismatch between the \gls{lut} and the dynamics of the real vehicle, the nominal LUT steering command may not produce the expected lateral acceleration. To compensate for this mismatch, MAP2 learns a steering residual from real-world driving data collected on the F1TENTH vehicle.

We construct the residual training label as the difference between the response-equivalent steering angle $\delta_{\mathrm{resp},k}$ and the nominal steering angle $\delta_{\mathrm{nom},k}$, both computed with the LUT lookup function $\mathcal{L}$:
\begin{align}
\delta_{\mathrm{resp},k} &= \mathcal{L}\left(a_y(t_k+\tau),\, v_x^{\mathrm{LUT}}(t_k),\, \widehat{F}_{x,f}(t_k+\tau)\right) \\
\delta_{\mathrm{nom},k}  &= \mathcal{L}\left(a_c^{*}(t_k),\, v_x^{\mathrm{LUT}}(t_k),\, \widehat{F}_{x,f}(t_k)\right) \\
\Delta\delta_k &= \delta_{\mathrm{resp},k} - \delta_{\mathrm{nom},k}
\end{align}
where $\mathcal{L}$ denotes the LUT lookup function; $a_c^{*}(t_k)$ is the lateral-acceleration request generated by the MPC at control step $k$; $v_x^{\mathrm{LUT}}(t_k)$ is the longitudinal velocity used as the LUT query input; $a_y$ and $a_x$ are the measured lateral and longitudinal accelerations; $\widehat{F}_{x,f}=\lambda_f m a_x$ is the front-axle longitudinal force estimated from the measured longitudinal acceleration, with $\lambda_f$ the front-axle drive fraction and $m$ the vehicle mass; $t_k$ denotes the sampling instant of control step $k$; and $\tau$ denotes the response delay, in seconds, estimated offline by cross-correlating the applied steering command $\delta_{\mathrm{cmd}}$ with $a_y$. Each training sample $k$ corresponds to one control cycle, and the measured signals at $t_k+\tau$ are evaluated by linear interpolation.

The SGP input feature vector is
\begin{align}
z_k = [v_{x,k}, r_k, a_{x,k}, a_{y,k}, \delta_{\mathrm{prev},k}, v_{y,k}, \beta_k]^{\mathsf T},
\end{align}
where $z_k$ denotes the SGP input feature vector; $v_{x,k}$ and $v_{y,k}$ are the measured longitudinal and lateral velocities; $r_k=\dot\Psi_k$ denotes the measured/estimated yaw rate; $a_{x,k}$ and $a_{y,k}$ are the measured longitudinal and lateral accelerations; $\delta_{\mathrm{prev},k}=\delta_{\mathrm{cmd},k-1}$ denotes the steering command applied in the previous control cycle; and $\beta_k$ denotes the vehicle sideslip angle.

The trained sparse GP predicts the steering-residual correction $\widehat{\Delta\delta}_k$, and the corrected steering command before limiting is computed as
\begin{align}
\delta_k^{\mathrm{pre}} = \delta_{\mathrm{nom},k} + \widehat{\Delta\delta}_k
\end{align}
where $\delta_k^{\mathrm{pre}}$ denotes the corrected command before applying the steering-magnitude and per-cycle steering-increment constraints. The final steering command is
\begin{align}
\delta_{\mathrm{cmd},k} = \mathrm{clip}\left(\delta_k^{\mathrm{pre}},\, -\delta_{\max},\, \delta_{\max}\right)
\end{align}
subject to $|\delta_{\mathrm{cmd},k}-\delta_{\mathrm{cmd},k-1}|\le\Delta\delta_{\max}$, where $\delta_{\max}$ and $\Delta\delta_{\max}$ are the steering-magnitude and per-cycle increment limits (Table~\ref{tab:ctrl_param}).

The residual model is implemented with GPyTorch~\cite{gardner2018gpytorch} using the \texttt{InducingPointKernel} wrapper, which realizes the SGPR inducing-variable approximation~\cite{titsias2009variational}. We use a constant mean and a scaled RBF kernel, a Gaussian likelihood, and $M{=}200$ inducing points initialized from the training inputs. Hyperparameters are optimized for $300$ Adam steps (learning rate $=0.1$) by maximizing the collapsed variational lower bound on the marginal log-likelihood. Inputs are standardized per feature and the target is scaled to $[-1,1]$.

\section{Experiments}

\subsection{Experimental Setup}
\begin{figure}[htbp]
  \centering
  \includegraphics[width=0.85\columnwidth]{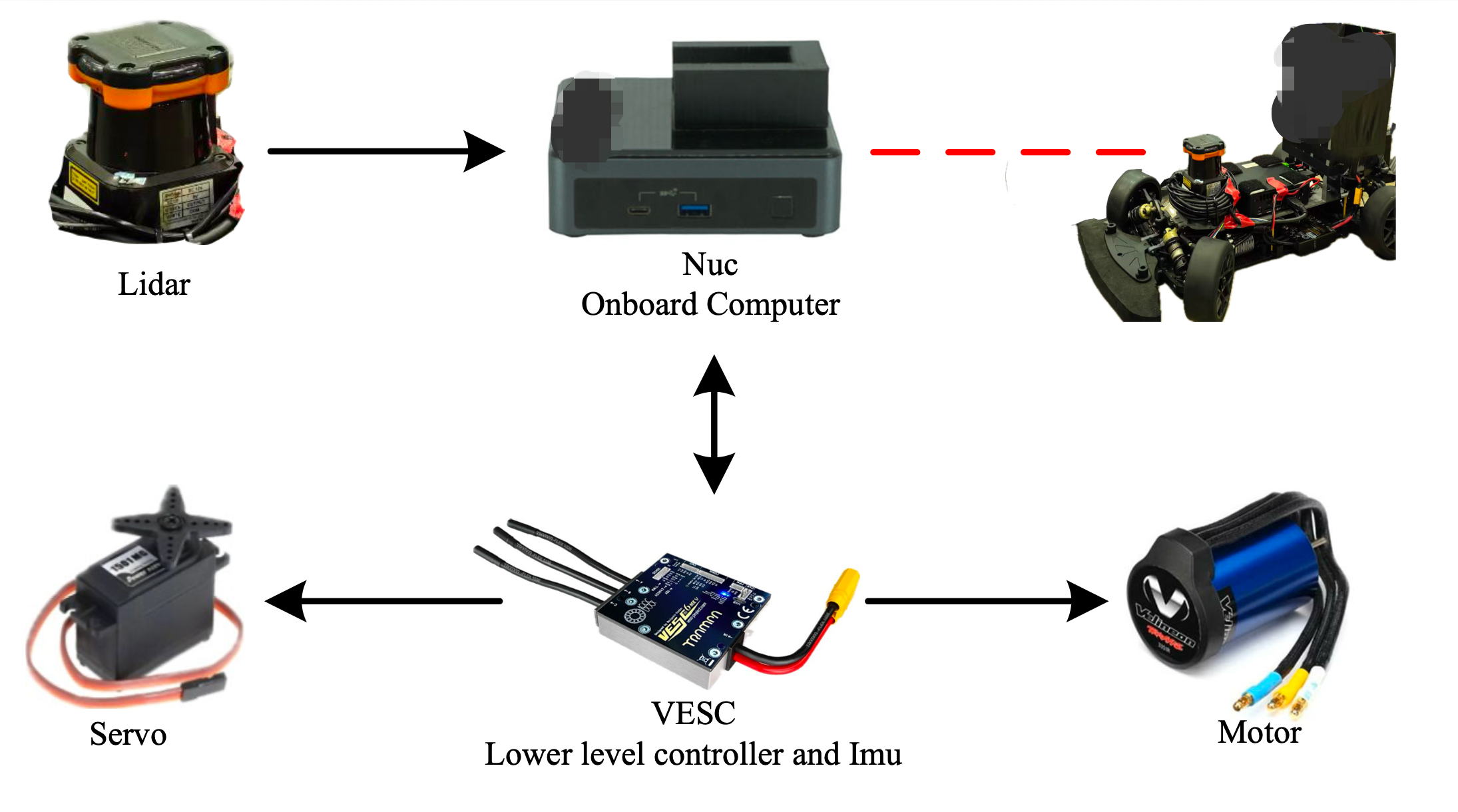}
  \caption{1:10-scale autonomous F1TENTH racing platform.}
  \label{fig:car_fig}
\end{figure}

Physical experiments were conducted using a 1:10-scale car. The hardware platform, shown in \cref{fig:car_fig}, is based on the Serpent SRX8GTE AWD car. The sensor suite includes a Hokuyo UST-30LX LiDAR with a maximum scan frequency of 40 Hz, and odometry is provided by the \gls{VESC}. Sensor fusion of data from LiDAR, IMU embedded in \gls{VESC}, and odometry gets the vehicle’s state. More details about localization are provided in~\cite{forzaeth}. An Intel NUC serves as the onboard computer, running the Robot Operating System. The optimization problem about MPC was solved by acados~\cite{acados}.

\begin{figure}[htbp]
  \centering
  \includegraphics[width=0.85\columnwidth]{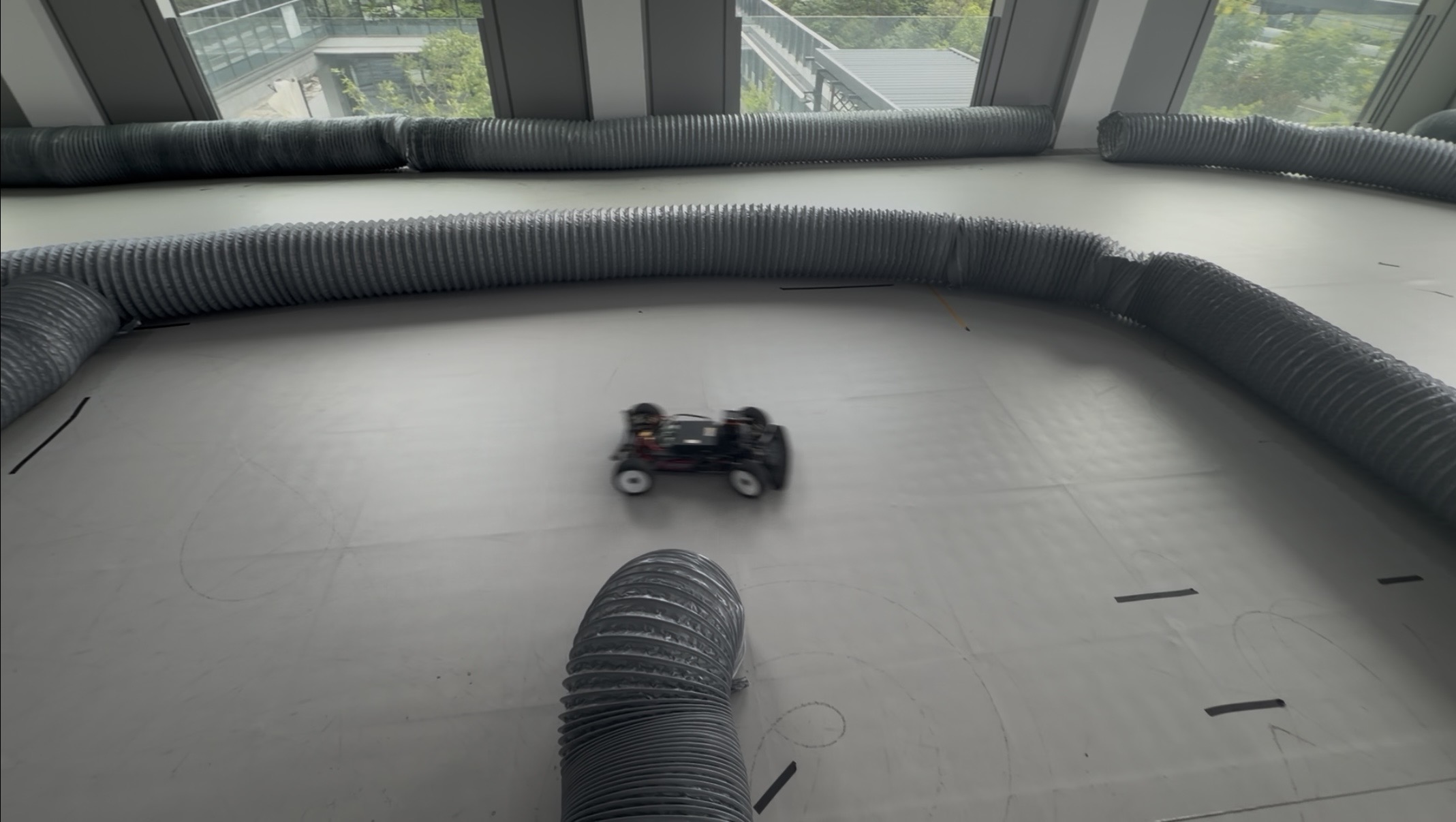}
  \caption{Real-world experiment.}
  \label{fig:real_car_fig}
\end{figure}

\begin{table}[ht]
        \centering
        \caption{Vehicle Parameters}\label{tab:vehicle_parm}
        \begin{tabular}{c|c|c|c}
        \toprule
        \textbf{ Parameters}& \textbf{Value}& \textbf{ Parameters}&\textbf{Value}\\
        \midrule
        $m$&           $4.24$ kg&                   $[B_f, B_r]$&           $[6.03, 18.67]$\\
        $I_{z}$&       $0.067$ kg$\cdot$m$^{2}$&     $[C_f, C_r]$&           $[1.26, 1.27]$\\
        $l_f$&         $0.168$ m&                    $[D_f, D_r]$&           $[0.87, 0.87]$\\
        $l_r$&         $0.162$ m&                    $[E_f, E_r]$&           $[0.19, 0.27]$\\
        $\mu $&        $1.0$&                    &                    \\
        \bottomrule
        \end{tabular}
\end{table}

\subsection{System Identification}
\begin{figure}[htbp]
  \centering
  \includegraphics[scale=0.35]{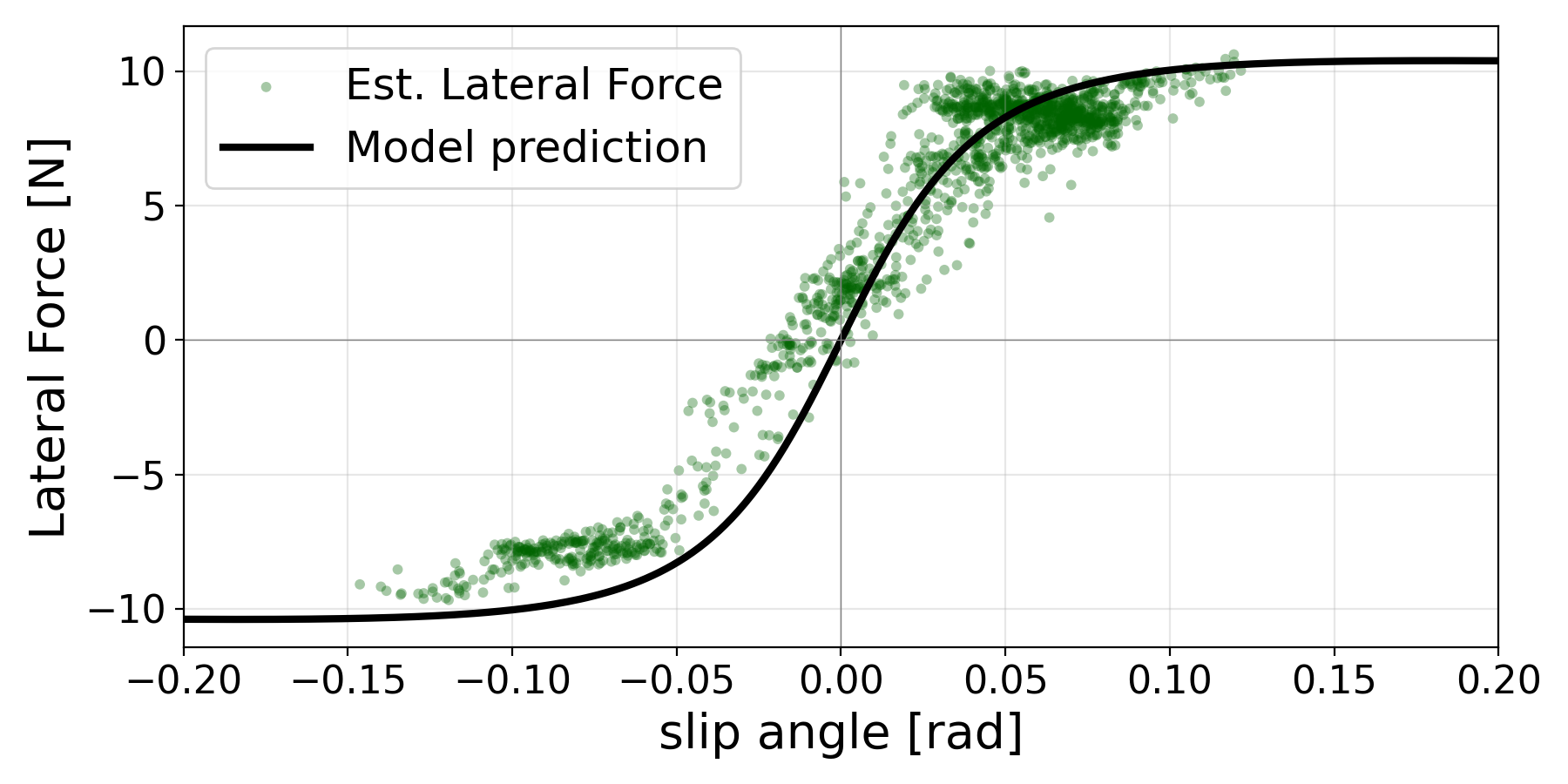}
  \caption{Data points obtained from the steady-state cornering experiment and the resulting model fit of the Pacejka model.
}
  \label{fig:pacejka_fit}
\end{figure}

Tire parameters were identified from steady-state cornering experiments following Voser et al.~\cite{voser2010analysis}. The steering angle was increased incrementally at approximately constant speed, and the resulting IMU measurements were used to estimate the axle lateral and longitudinal forces and fit the Pacejka model. The resulting force-slip relationship is shown in Fig.~\ref{fig:pacejka_fit}.

\begin{align}
F_{y,f}&=\frac{m\,l_r\,a_{y,imu}-(l_f+l_r)\,F_{x,f}\sin\delta}{(l_f+l_r)\,\cos\delta}\\
F_{y,r}&=\frac{m\,l_f}{l_f+l_r}\,a_{y,imu}
\end{align}

The specific parameters are listed in \cref{tab:vehicle_parm}. To better identify the peak factor, a dynamometer measurement was performed, and the peak factor was set to 0.87.
The control parameters are listed in \cref{tab:ctrl_param}.

\begin{table}[ht]
        \centering
            \renewcommand{\arraystretch}{0.85}  
        \caption{MAP2 Control Parameters}\label{tab:ctrl_param}
        \begin{tabular}{c|c|c|c}
        \toprule
        \textbf{Parameters} & \textbf{Value} & \textbf{Parameters} & \textbf{Value}\\
        \midrule
        $N$&                 $64$&                         $\Delta t$&         $0.05\,\mathrm{s}$\\
        $q_n$&               $100.0$&                      $q_{e_\psi}$&   $30.0$\\
        $q_v$&               $14.0$&                       $r_{v_y}$&          $2.0$\\
        $r_{\dot\Psi}$&      $3.0$&                        $p_n$&              $300.0$\\
        $p_{e_\psi}$&    $50.0$&                       $\rho$&             $500.0$\\
        $v_{x,\min}$&        $0.5\,\mathrm{m/s}$&          $v_{x,\max}$&       $12.0\,\mathrm{m/s}$\\
        $v_{y,\max}$&        $0.05\,\mathrm{m/s}$&         $\dot\Psi_{\max}$&  $6.0\,\mathrm{rad/s}$\\
        $\delta_{\max}$&     $0.4\,\mathrm{rad}$&           $\Delta\delta_{\max}$& $0.4\,\mathrm{rad}$\\
        \bottomrule
        \end{tabular}

\end{table}

\subsection{Objective and Test Procedure}
\begin{figure}[htbp]
  \centering
  \begin{subfigure}[b]{0.48\linewidth}
        \centering
        \includegraphics[width=\linewidth]{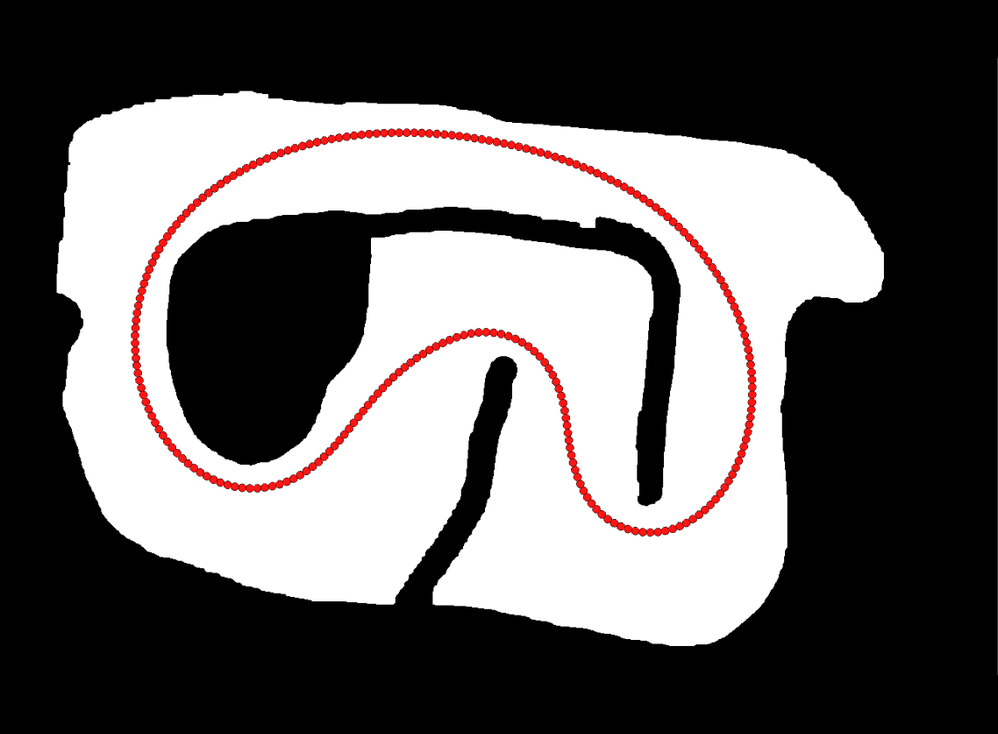}
        \caption{Map I}
        \label{fig:mapA}
    \end{subfigure}
  \begin{subfigure}[b]{0.48\linewidth}
        \centering
        \includegraphics[width=\linewidth]{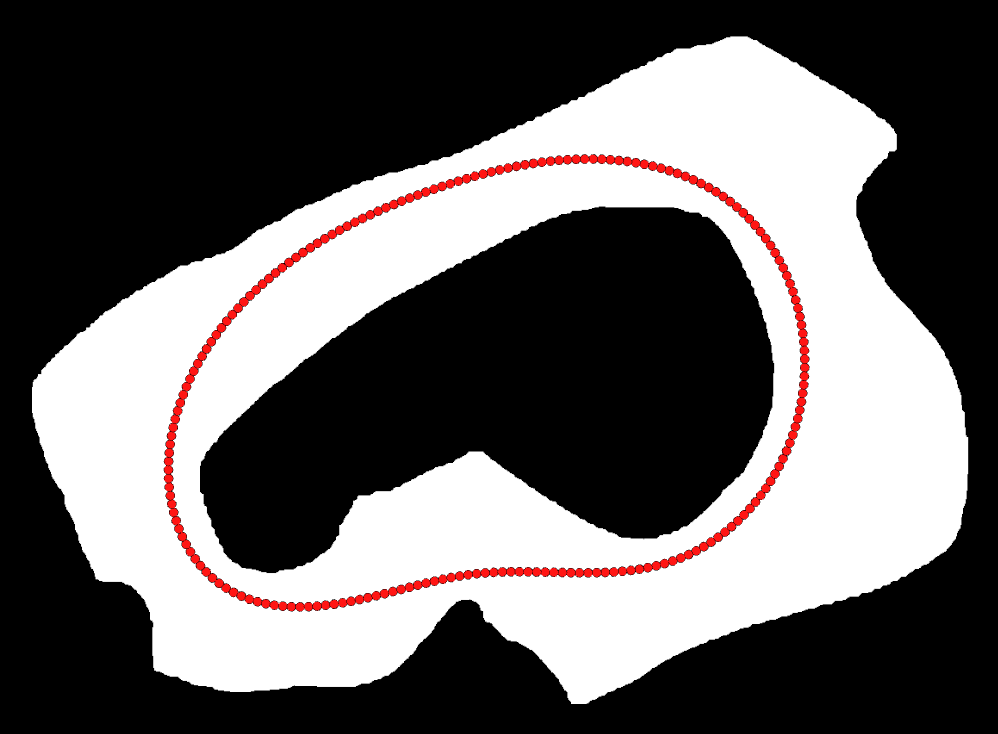}
        \caption{Map II}
        \label{fig:mapB}
    \end{subfigure}
  \caption{The tracks of two different maps.}
  \label{fig:maps}
\end{figure}

Our objective is to compare the tracking performance and lap times of the PP controller, MAP controller, and MAP2 controller with or without SGP in two different maps, which are shown in Fig.~\ref{fig:maps}. Additionally, from a practical perspective, the robustness of the controllers' performance is evaluated when the tire parameters are mismatched.

The global trajectory optimizer from ~\cite{raceline} was used to generate a minimum curvature trajectory and calculate the speed profile. Since the velocity profile is computed for perfect tracking, the experiment started at  50\% of the ideal speed and gradually increased until the vehicle crashes into the boundaries. For each test, the car will try to complete ten laps.

For the parameter sensitivity experiments, we performed a cross-tire validation: the tire model parameters identified on tire A were directly applied to tire B without re-identification, thereby exposing the algorithm to a realistic parameter mismatch. Specifically, for all three controllers, the offline LUT is generated using tire A's parameters only. MAP2 with SGP further retrains the SGP on driving data logged on tire B. We have defined four metrics to evaluate the sensitivity of the controller to changes in tire parameters. These include the  \gls{ALT}, \gls{FLT}, \gls{ALE}, and \gls{MLE}.

\subsection{Experiment Results}
\begin{figure}[htbp]
  \centering
  \begin{subfigure}[b]{0.48\linewidth}
        \centering
        \includegraphics[width=\linewidth]{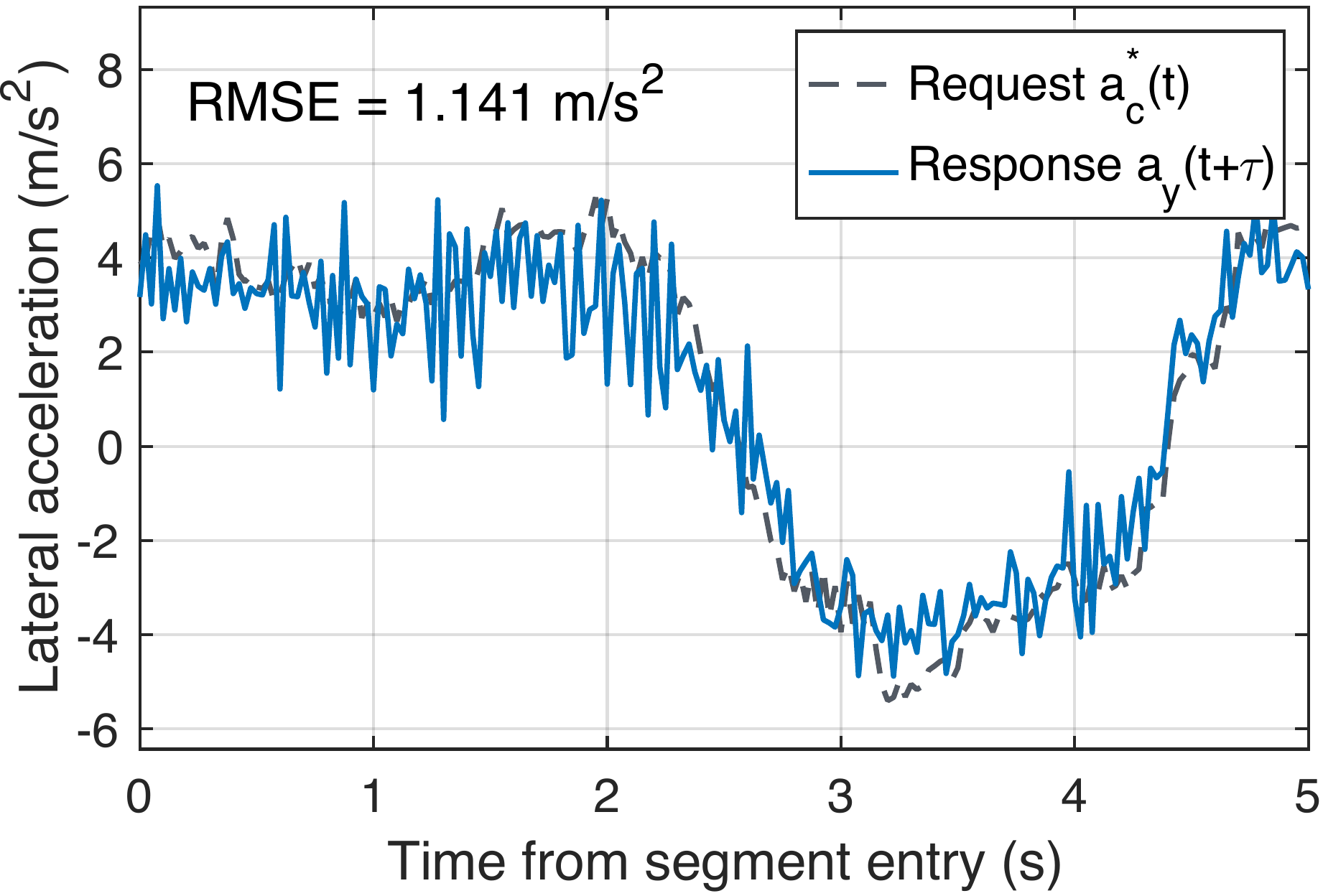}
        \caption{MAP2 with SGP}
        \label{fig:MAP2_with_SGP}
    \end{subfigure}
  \begin{subfigure}[b]{0.48\linewidth}
        \centering
        \includegraphics[width=\linewidth]{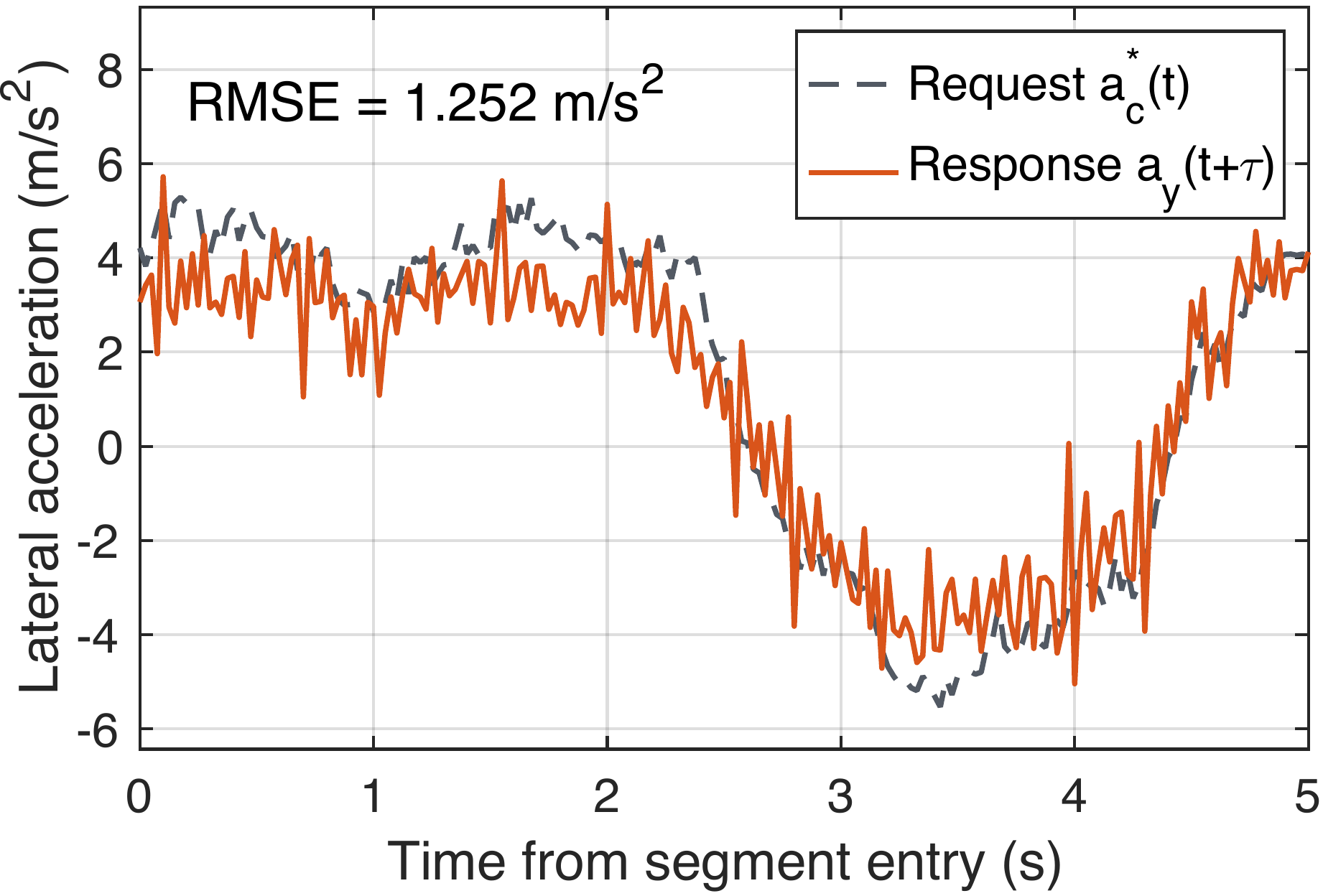}
        \caption{MAP2 without SGP}
        \label{fig:MAP2_without_SGP}
    \end{subfigure}
  \caption{Effect of SGP on lateral-acceleration tracking for a representative MAP2 segment.}
  \label{fig:sgp_improve}
\end{figure}

The results of average lateral error, max lateral error and lap time are reported in \autoref{tab:phys_ex_map1} and \autoref{tab:phys_ex_map2}, which show that the performance gap between the controllers becomes more apparent as the speed setting increases. At the highest speed of 75\%, Table~\ref{tab:phys_ex_map1} shows that, compared with the baseline \gls{map} controller, MAP2 reduces the average lateral tracking error by \textbf{37.99\%}, the maximum lateral tracking error by \textbf{57.48\%}, and the lap time by \textbf{1.57\%}. Compared with the classical PP controller, according to Table~\ref{tab:phys_ex_map2}, the average and maximum lateral tracking errors are reduced by \textbf{44.65\%} and \textbf{47.81\%}, respectively, together with a \textbf{3.27\%} reduction in lap time. While at low speed, the geometric controller PP and MAP can achieve tracking errors comparable to or lower than MAP2 with or without SGP, however, in practice the absolute lateral tracking error at low speed is in fact very small. Specifically, the additional SGP residual correction further improves high-speed MAP2 performance. As shown in Fig.~\ref{fig:sgp_improve}, SGP correction reduces the lateral-acceleration tracking RMSE from 1.252 to 1.141~$\mathrm{m/s^2}$ for the representative segments shown, corresponding to a reduction of \textbf{8.87\%}. This indicates closer agreement between the measured and requested lateral accelerations. The ablation results in Table~\ref{tab:phys_ex_map2} also show that, at the 75\% speed setting, SGP correction reduces the average lateral error, maximum lateral error, and average lap time by \textbf{26.15\%}, \textbf{22.16\%}, and \textbf{1.13\%}, respectively, compared with MAP2 without SGP.

\begin{figure}[htbp]
    \centering
    \begin{subfigure}[b]{0.31\linewidth}
        \centering
        \includegraphics[width=\linewidth]{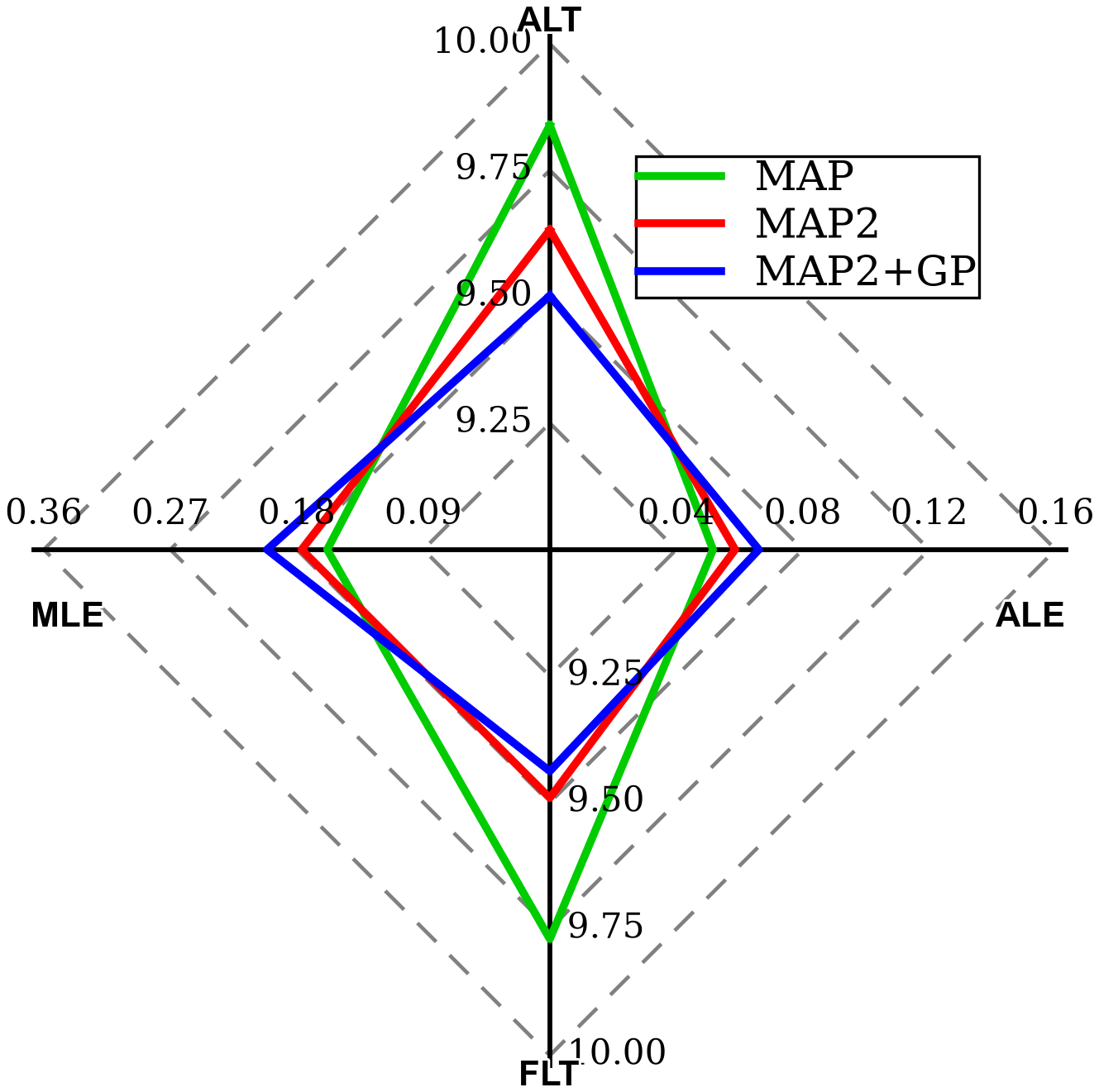}
        \caption{60\% Velocity }
        \label{fig:image1}
    \end{subfigure}
    \hfill
    \begin{subfigure}[b]{0.31\linewidth}
        \centering
        \includegraphics[width=\linewidth]{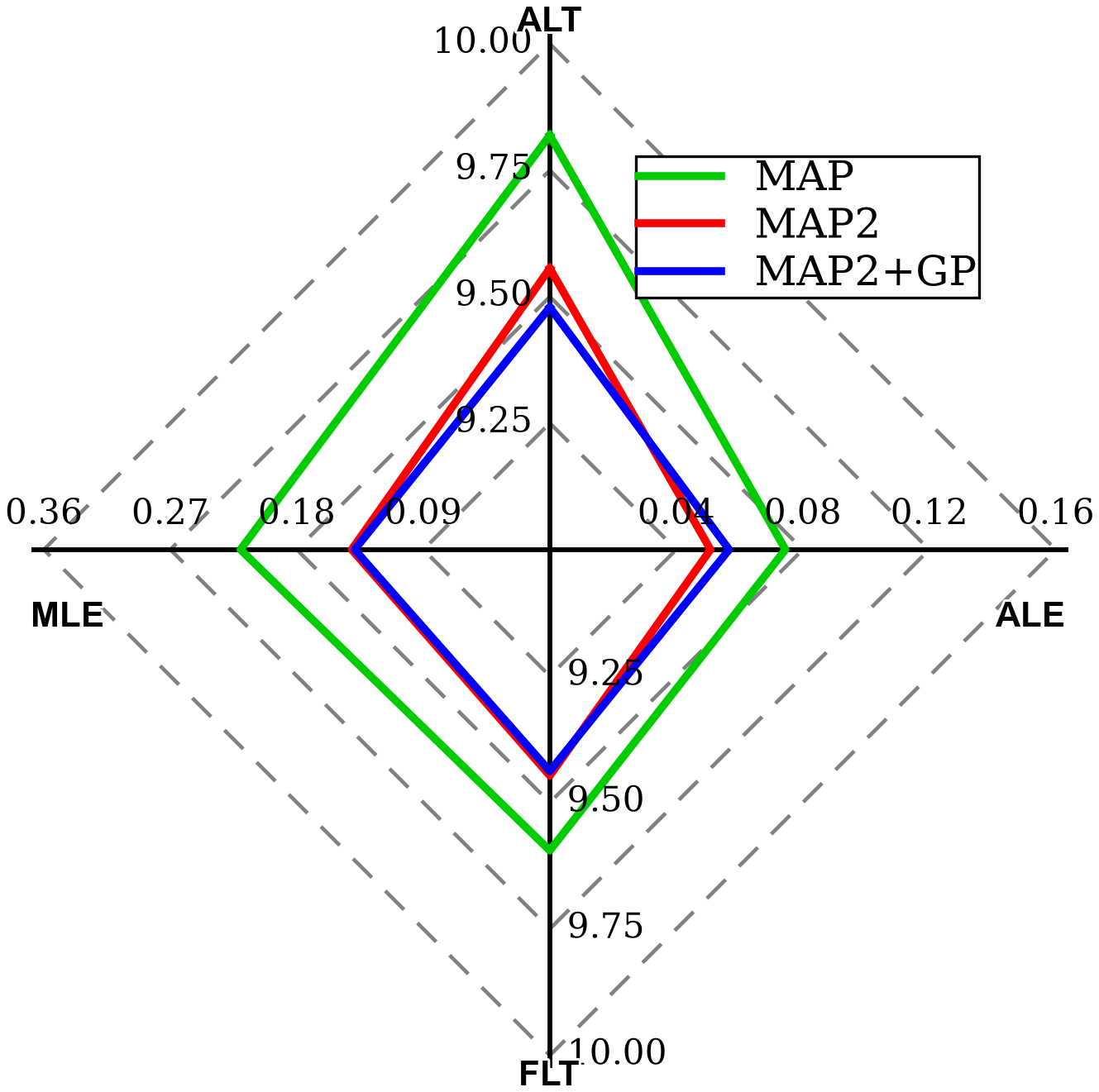}
        \caption{70\% Velocity }
        \label{fig:image2}
    \end{subfigure}
    \hfill
    \begin{subfigure}[b]{0.31\linewidth}
        \centering
        \includegraphics[width=\linewidth]{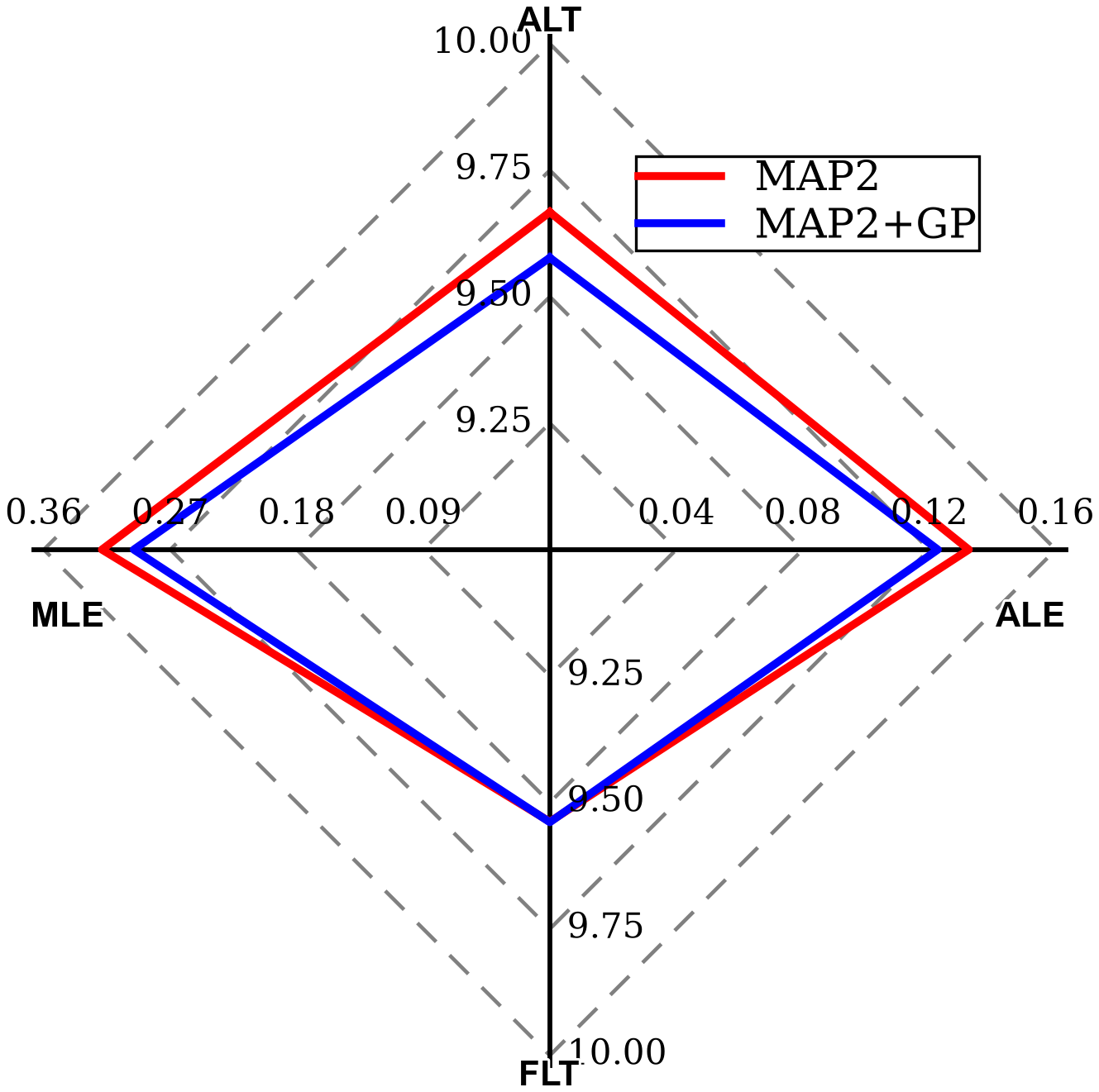}
        \caption{75\% Velocity }
        \label{fig:image3}
    \end{subfigure}
    \caption{The comparison of the controllers when tires and tire parameters mismatch (MAP in 75\% velocity crashed into the wall and did not finish)}
    \label{fig:param_sen}
\end{figure}

The cross-tire experiments further show the importance of residual correction when the tire parameters used to generate the offline LUT no longer accurately represent the actual vehicle dynamics. According to Fig.~\ref{fig:param_sen}, MAP2 substantially improves tracking performance over the original \gls{map} controller under this model mismatch, reducing the average and maximum lateral tracking errors by \textbf{23.89\%} and \textbf{36.89\%} at the velocity of 70\%, respectively, and at the velocity of 75\%, MAP2 finished 10 laps with good lateral error control, where \gls{map} crashed into the wall. The SGP correction provides additional improvements without requiring regeneration of the nominal LUT, demonstrating that the combination of a physics-based baseline and data-driven residual learning can effectively compensate for modeling inaccuracies. Meanwhile, the complete MAP2 control cycle requires approximately \textbf{9~ms} on the onboard computing platform, demonstrating its suitability for real-time autonomous racing on computationally constrained systems.

\begin{table*}[t]
\caption{Controller performance on the \textbf{physical system} in map I}
    \centering
    \renewcommand{\arraystretch}{0.85}    
    \begin{tabular}{c|c||c|c||c|c||c|c|c}
    \toprule
    \multirow{2}{*}{\textbf{Velocity scale (\%)}} 
    &\multirow{2}{*}{\textbf{Controller}} 
    &\multicolumn{2}{c||}{\textbf{Avg. Lateral error (m)}} 
    & \multicolumn{2}{c||}{\textbf{Max. Lateral error (m)}} 
    & \multicolumn{3}{c}{\textbf{Lap time (s)}}      \\
    \cmidrule(l{5pt}r{5pt}){3-4}
    \cmidrule(l{5pt}r{5pt}){5-6}
    \cmidrule(l{5pt}r{5pt}){7-9}
    &
    &\textbf{Avg.}& \textbf{Std.} 
    &\textbf{Avg.}& \textbf{Std.}   
    &\textbf{Fastest}  &\textbf{Avg.}  &\textbf{Std.}\\
    \midrule
    \multirow{4}{*}{\textbf{50}}    
    & PP        & \textbf{0.0298} & 0.0007 & 0.1035 & 0.0085 & 11.054 & 11.097 & 0.021 \\
    & MAP       & 0.0325 & 0.0018 &\textbf{0.0851} & 0.0030 & 10.946 & \textbf{10.981} & 0.034 \\
    & MAP2 w/o SGP & 0.0471 & 0.0022 & 0.1076 & 0.0078 & 11.070 & 11.121 & 0.034 \\
    & MAP2 w/ SGP  & 0.0504 & 0.0028 & 0.1190 & 0.0087 & \textbf{10.942} & 10.997 & 0.031 \\
    \midrule[0.5\lightrulewidth]
\multirow{4}{*}{\textbf{60}}   
    & PP              & 0.0723        & 0.0015 & 0.2100        & 0.0171 & 9.582          & 9.600          & 0.014 \\
    & MAP             & 0.0561        & 0.0057 & 0.1757        & 0.0275 & 9.490          & 9.548          & 0.037 \\
    & MAP2 w/o SGP     & 0.0598 & 0.0028 & 0.1574 & 0.0128 & 9.478 & 9.540 & 0.037 \\
    & MAP2 w/ SGP      & \textbf{0.0508} & 0.0037 & \textbf{0.1545} & 0.0171 & \textbf{9.322} & \textbf{9.430} & 0.052 \\
\midrule[0.5\lightrulewidth]
\multirow{4}{*}{\textbf{70}}   
    & PP              & 0.2147        & 0.0182 & 0.5006        & 0.0392 & 9.522          & 9.643          & 0.067 \\
    & MAP             & 0.1764        & 0.0213 & 0.5303        & 0.0975 & 9.342          & 9.507          & 0.123 \\
    & MAP2 w/o SGP     & 0.1304 & 0.0080 & 0.2715 & 0.0334 & 9.115 & 9.210 & 0.051 \\
    & MAP2 w/ SGP      & \textbf{0.1059} & 0.0107 & \textbf{0.2439} & 0.0377 & \textbf{8.966} & \textbf{9.061} & 0.063 \\
\midrule[0.5\lightrulewidth]
\multirow{4}{*}{\textbf{75}}  
    & PP              & 0.1992        & 0.0096 & 0.4632        & 0.0480 & 9.218          & 9.314          & 0.083 \\
    & MAP             & 0.2140        & 0.0263 & 0.7291        & 0.0886 & 9.110          & 9.252          & 0.134 \\
    & MAP2 w/o SGP     & 0.1433 & 0.0111 & 0.3404 & 0.0476 & 9.043 & 9.174 & 0.083 \\
    & MAP2 w/ SGP      & \textbf{0.1327} & 0.0106 & \textbf{0.3100} & 0.0385 & \textbf{8.982} & \textbf{9.107} & 0.083 \\
    \bottomrule
    \end{tabular}%
    
    \label{tab:phys_ex_map1}
\end{table*}

\begin{table*}[t]
\caption{Controller performance on the \textbf{physical system} in map II}
\centering
    \renewcommand{\arraystretch}{0.85}  
\begin{tabular}{c|c||c|c||c|c||c|c|c}
\toprule
\multirow{2}{*}{\textbf{Velocity scale (\%)}}
& \multirow{2}{*}{\textbf{Controller}}
& \multicolumn{2}{c||}{\textbf{Avg. Lateral error (m)}}
& \multicolumn{2}{c||}{\textbf{Max. Lateral error (m)}}
& \multicolumn{3}{c}{\textbf{Lap time (s)}} \\
\cmidrule(l{5pt}r{5pt}){3-4}
\cmidrule(l{5pt}r{5pt}){5-6}
\cmidrule(l{5pt}r{5pt}){7-9}
&
& \textbf{Avg.} & \textbf{Std.}
& \textbf{Avg.} & \textbf{Std.}
& \textbf{Fastest} & \textbf{Avg.} & \textbf{Std.} \\
\midrule
\multirow{4}{*}{\textbf{50}}
& PP              & 0.0466 & 0.0013 & 0.1655 & 0.0063 & 8.146 & 8.172 & 0.018 \\
& MAP             & \textbf{0.0243} & 0.0008 & 0.1140 & 0.0063 & 8.054 & \textbf{8.080} & 0.019 \\
& MAP2 w/o SGP    & 0.0503 & 0.0022 & 0.1088 & 0.0064 & 8.159 & 8.203 & 0.032 \\
& MAP2 w/ SGP     & 0.0371 & 0.0018 & \textbf{0.1056} & 0.0097 & \textbf{8.047} & 8.099 & 0.038 \\
\midrule[0.5\lightrulewidth]
\multirow{4}{*}{\textbf{60}}
& PP              & 0.1327 & 0.0029 & 0.3584 & 0.0160 & 7.031 & 7.087 & 0.033 \\
& MAP             & \textbf{0.0638} & 0.0045 & 0.1801 & 0.0137 & 6.863 & 6.910 & 0.031 \\
& MAP2 w/o SGP    & 0.0961 & 0.0039 & 0.1787 & 0.0217 & 6.926 & 6.979 & 0.030 \\
& MAP2 w/ SGP     & 0.0749 & 0.0028 & \textbf{0.1721} & 0.0182 & \textbf{6.826} & \textbf{6.904} & 0.038 \\
\midrule[0.5\lightrulewidth]
\multirow{4}{*}{\textbf{70}}
& PP              & 0.2142 & 0.0061 & 0.5793 & 0.0310 & 7.042 & 7.091 & 0.028 \\
& MAP             & 0.2382 & 0.0249 & 0.8210 & 0.0666 & 7.091 & 7.167 & 0.062 \\
& MAP2 w/o SGP    & 0.1181 & 0.0054 & 0.1922 & 0.0187 & 6.763 & 6.842 & 0.041 \\
& MAP2 w/ SGP     & \textbf{0.0979} & 0.0055 & \textbf{0.1787} & 0.0159 & \textbf{6.756} & \textbf{6.806} & 0.031 \\
\midrule[0.5\lightrulewidth]
\multirow{4}{*}{\textbf{75}}
& PP              & 0.1908 & 0.0112 & 0.4388 & 0.0425 & 6.918 & 6.983 & 0.047 \\
& MAP             & 0.2766 & 0.0338 & 0.9196 & 0.1659 & - & $^{\dagger}$ & - \\
& MAP2 w/o SGP    & 0.1430 & 0.0066 & 0.2942 & 0.0308 & 6.766 & 6.832 & 0.034 \\
& MAP2 w/ SGP     & \textbf{0.1056} & 0.0079 & \textbf{0.2290} & 0.0151 & \textbf{6.691} & \textbf{6.755} & 0.057 \\
\bottomrule
\multicolumn{9}{l}{\small $\dagger$\ MAP crashed into the wall and could not finish 10 laps.} \\
\end{tabular}%
\label{tab:phys_ex_map2}
\end{table*}

\section{Conclusions}
\label{conclusion}
In this paper, we proposed MAP2, a computationally lightweight and robust lateral controller for high-performance autonomous racing. The proposed framework combines a kinematic \gls{mpc} for predictive lateral-acceleration generation, a model-based \gls{lut} for steering computation, and an \gls{sgp} for residual correction. This combination improves predictive tracking capability and compensates for model mismatch while maintaining low computational latency.

The proposed controller was experimentally validated on a 1:10-scale F1TENTH platform on two different tracks and over multiple speed settings. The results demonstrate reduced lateral tracking errors and lap times at high speeds relative to \gls{map} and \gls{pp}, with additional improvements from SGP residual correction. The cross-tire experiments further demonstrate robustness to tire-parameter mismatch, with SGP retraining providing compensation without regenerating the nominal LUT. Together with its low computational cost, these results demonstrate the suitability of MAP2 for real-time autonomous racing on computationally constrained systems.

Future work will focus on online adaptation of the SGP inducing points for the MAP2 controller such that the residual model can be updated incrementally at runtime without offline retraining when operating conditions change. Furthermore, testing the behavior of the validated MAP2 controller on a full-scale racing car would be of great interest.





\section*{ACKNOWLEDGMENT}
The authors would like to thank all members of the DDRX team.

\bibliographystyle{IEEEtran}
\bibliography{main}

\end{document}